\documentclass[letterpaper, 10 pt, conference]{ieeeconf}  

\IEEEoverridecommandlockouts                              

\usepackage{graphicx}
\usepackage{float}  
\usepackage{subfigure}  
\usepackage{multicol}
\usepackage{amsmath}
\usepackage{amssymb}
\usepackage{soul}
\usepackage{hyperref}
\usepackage{color}

\usepackage{booktabs}
\usepackage{multirow}
\usepackage[ruled,linesnumbered,noend]{algorithm2e} 

\title{\LARGE \bf
Local Path Optimization in The Latent Space Using Learned Distance Gradient
}

\author{Jiawei Zhang$^{1}$, Chengchao Bai$^{1}$, Wei Pan$^{2}$, Tianhang Liu$^{1}$ and Jifeng Guo$^{1}$
\thanks{}
\thanks{$^{1}$Jiawei Zhang, Chengchao Bai, Tianhang Liu and Jifeng Guo are with Harbin Institute of Technology, China. For correspondence: {\tt\small baichengchao@hit.edu.cn}}%
\thanks{$^{2}$ Wei Pan is with The Univeristy of Manchester, UK.}
}

\begin{document}

\maketitle
\thispagestyle{empty}
\pagestyle{empty}

\begin{abstract}

Constrained motion planning is a common but challenging problem in robotic manipulation. In recent years, data-driven constrained motion planning algorithms have shown impressive planning speed and success rate. Among them, the latent motion method based on manifold approximation is the most efficient planning algorithm. Due to errors in manifold approximation and the difficulty in accurately identifying collision conflicts within the latent space, time-consuming path validity checks and path replanning are required. In this paper, we propose a method that trains a neural network to predict the minimum distance between the robot and obstacles using latent vectors as inputs. The learned distance gradient is then used to calculate the direction of movement in the latent space to move the robot away from obstacles. Based on this, a local path optimization algorithm in the latent space is proposed, and it is integrated with the path validity checking process to reduce the time of replanning. The proposed method is compared with state-of-the-art algorithms in multiple planning scenarios, demonstrating the fastest planning speed.

\end{abstract}

\section{INTRODUCTION}

Motion constraints appear in many robotic tasks, such as the end-effector orientation constraint when the robot grasps a cup, and the closed-chain constraint in multi-robot cooperative manipulation \cite{c1}. Constrained motion planning is to calculate the collision-free motion path from the starting configuration to the goal configuration while satisfying the constraints. Compared to unconstrained motion planning problems, constraint-based motion planning is more complex. Sampling-based motion planning algorithms are currently the primary approach for constrained motion planning \cite{c2,c3}, which have the advantages of probabilistic completeness and do not need to explicitly parameterize the constraint manifold.

A key challenge in sampling-based motion planning is how to quickly obtain configurations that satisfy the constraints. Projection-based methods provide a simple and effective solution by projecting any configuration onto the constrained manifold using random gradient descent \cite{c4} or the Jacobian matrix of constraint functions \cite{c5,c6}. Based on this, the planning time can be further reduced by incorporating local linear approximations to the constraint manifold, such as Atlas RRT \cite{c7} and Tangent Bundle RRT \cite{c8}. However, in complex and high-dimensional constrained motion planning problems, these methods remain time-consuming.

   \begin{figure}[t]
      \centering
      \includegraphics[width=0.48\textwidth]{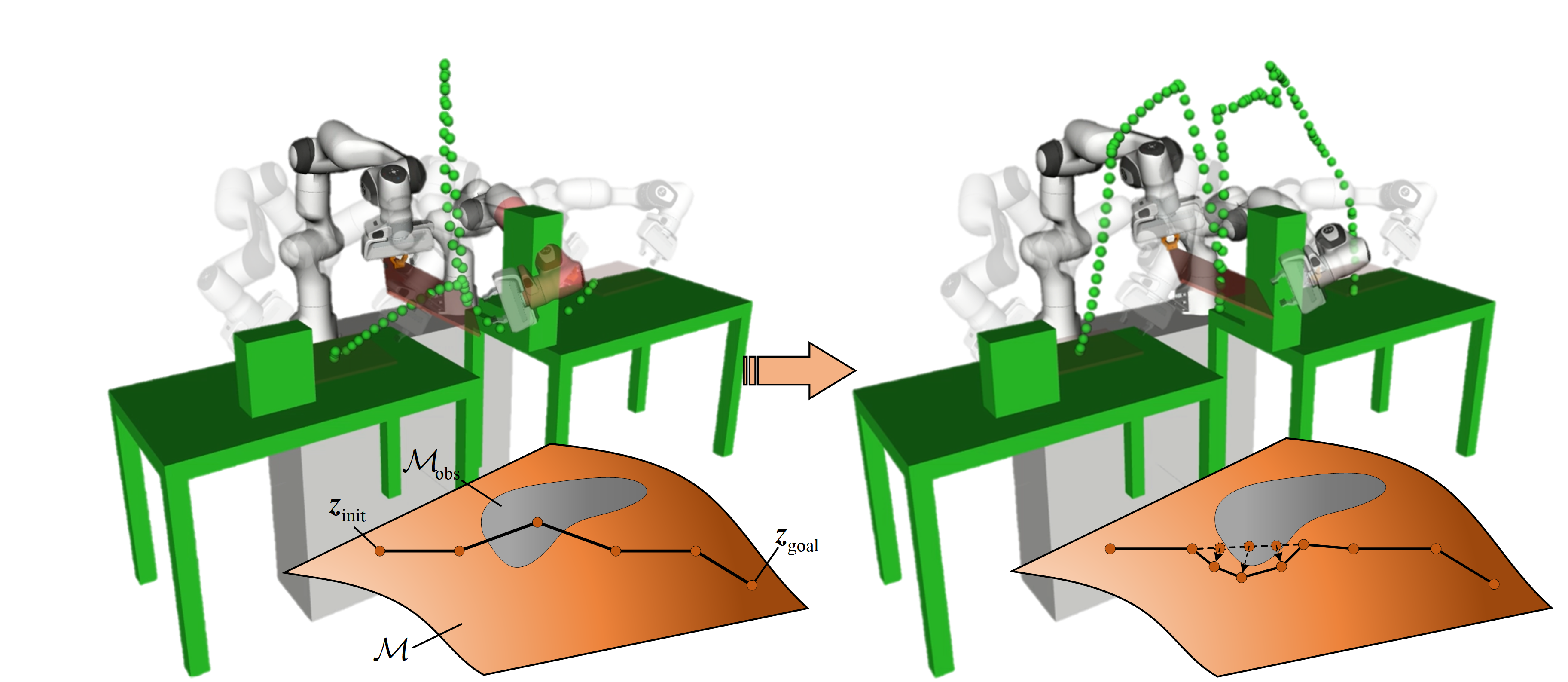} 
      \caption{Schematic of the proposed method, by performing local path optimization in the latent space, the robot gets out of the obstacle area.}
      \label{fig1}
      \label{figurelabel}
   \end{figure}

In recent years, data-driven constrained motion planning algorithms have shown obvious advantages in planning efficiency. These methods reduce planning time by precom
puted graph \cite{c9}, learning sampling distributions \cite{c10,c11}, or approximating constraint manifolds \cite{c12}. Manifold approximation methods learning the underlying distribution of manifold data and mapping it into a low-dimensional latent space, which can be regarded as a method to reparameterize the configuration space. By leveraging the latent space’s capability for sampling and continuous interpolation, random trees can be expanded directly in the latent space, which achieves the fastest planning speed at present \cite{c13}.

However, due to the manifold approximation error of neural networks and the difficulty in accurately identifying collision conflicts within the latent space, paths planned in the latent space may be invalid when mapped back to the configuration space. Therefore, the algorithm needs to check the validity of the latent path, which leads to the main time consumption. In this paper, a local path optimization algorithm in latent space is proposed, which directly moves latent waypoint away from obstacles area in latent space, successfully reduce the time-consuming of path replanning, the process is illustrated Fig.~\ref{fig1}.

\section{RELATED WORK}

\subsection{Data-free Algorithms}

Current research on constrained motion planning problems mainly focuses on sampling-based algorithms. Constraint relaxation \cite{c14,c15} is a class of simple and straightforward methods that make the probability of randomly sampling a configuration that satisfies the constraints no longer zero by allowing some error in the motion constraints. Constraint relaxation allows constrained motion planning problems to be solved directly using unconstrained sampling motion planning algorithms, but such methods are time-consuming due to the low probability of sampling to a feasible configuration.

Projection-based algorithms are more efficient and can be divided into inverse kinematics based methods and numerical based methods. Inverse kinematics based methods are mainly used for motion planning problems with closed-chain constraints, which require splitting the robot system into active and passive chains, first sampling the configurations of the active chain randomly, and then using inverse kinematics algorithms to compute the configurations of the passive chain that satisfy the constraints \cite{c16,c17,c18}. Numerical based projection methods can be used for all constraints, which can be divided into stochastic gradient descent methods \cite{c4} and Newton-Raphson projection methods  \cite{c5,c6}. The Newton-Raphson projection methods use the pseudo-inverse of the constrained Jacobian matrix to quickly calculate the direction of movement toward the constrained manifold, which have been widely used in the constrained motion planning problem.

In the projection-based algorithms, a large number of projection operations will increase the consumption of time. In order to reduce the planning time, the constraint manifold can be locally linearly approximated by the tangent space of the manifold. Tangent-space-based methods include Atlas RRT \cite{c7} and Tangent-Bundle RRT \cite{c8}. In addition, the configuration space can be reparameterized to obtain a new state space satisfying the constraints, and then various sampling-based motion planning algorithms can be directly applied in this reparameterized state space \cite{c19,c20}. Based on the above methods, Z. Kingston et al. proposed the framework of implicit space representation \cite{c21}, which can decouple the sampling-based planner and the methods for simulating constraints, and the flexible combination can be used to select the optimal planning scheme for specific tasks.

\subsection{Data-driven Algorithms}

Data-driven motion planning algorithms have been widely studied in unconstrained motion planning problems, which have the advantage of high planning speed \cite{c22}. For constrained motion planning problems, the Precomputed Graph[9] method uses the off-line precomputed graph to approximate the constraint manifold. It reduces the planning time by sampling the nodes in the precomputed graph during the motion planning process. When the constraint parameters change, the graph needs to be recomputed, which leads to poor flexibility of the algorithm. The CoMPNetX \cite{c10,c11} method uses precomputed motion paths to train a neural network for sampling, which speeds up the search process by sampling directionally. Such methods require a large amount of offline data, and the effect of the sampling network is affected by the quality of the precomputed paths. Manifold approximation methods map manifold to a low-dimensional latent space by learning the underlying distribution of manifold data. Then, a large number of configurations near the constraint manifold can be quickly generated by decoding the sampled latent vectors into the configuration space  \cite{c12}. In the latest study, the random tree is directly expanded in the latent space to generate latent paths, which achieves the fastest planning speed \cite{c13}.

   \begin{figure}[t]
      \centering
      \includegraphics[width=0.48\textwidth]{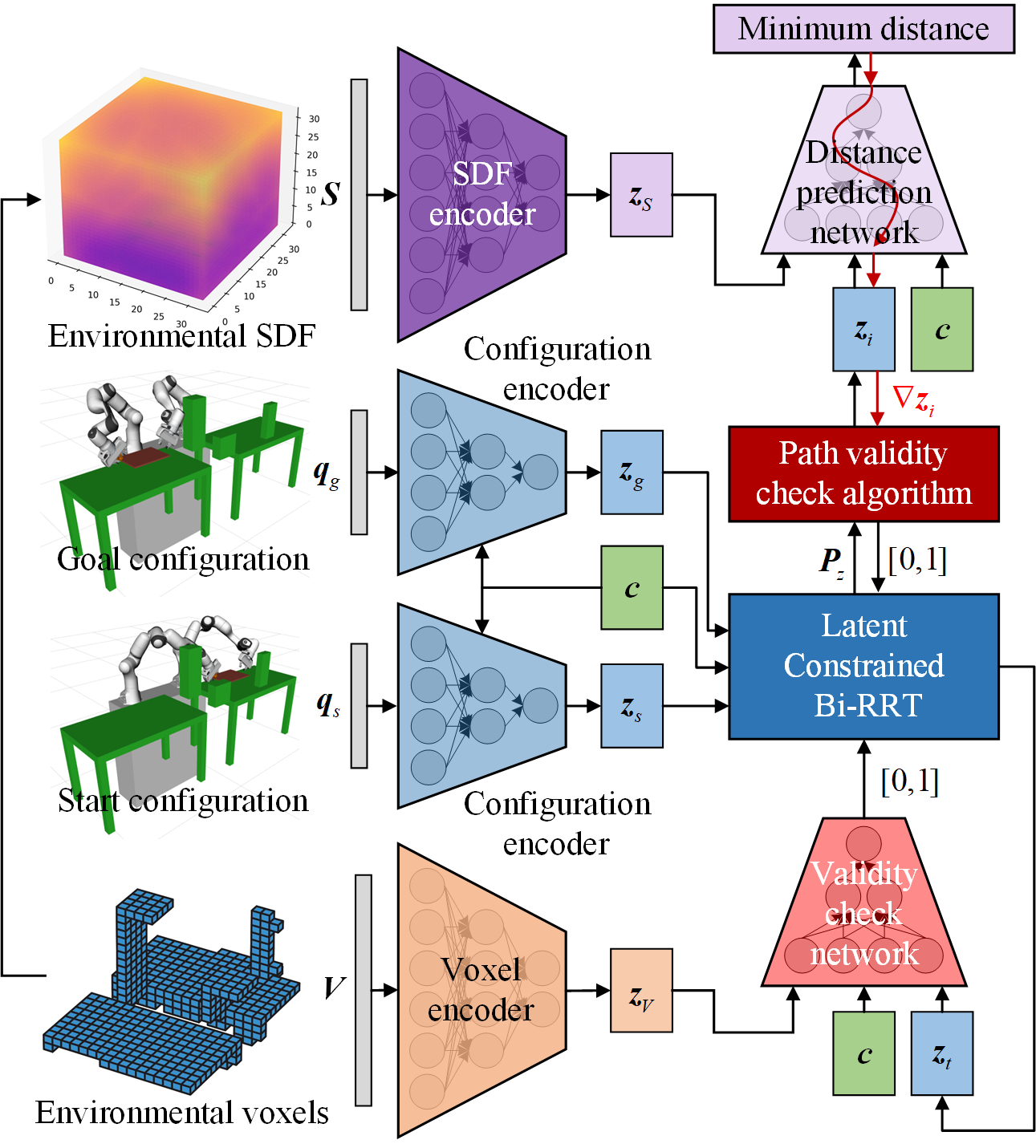} 
      \caption{The workflow of the proposed method.}
      \label{fig2}
      \label{figurelabel}
   \end{figure}

\section{PRELIMINARIES}

\subsection{Constrained Motion Planning}

\begin{figure*}[thpb]
  \centering
  \includegraphics[width=0.9\textwidth]{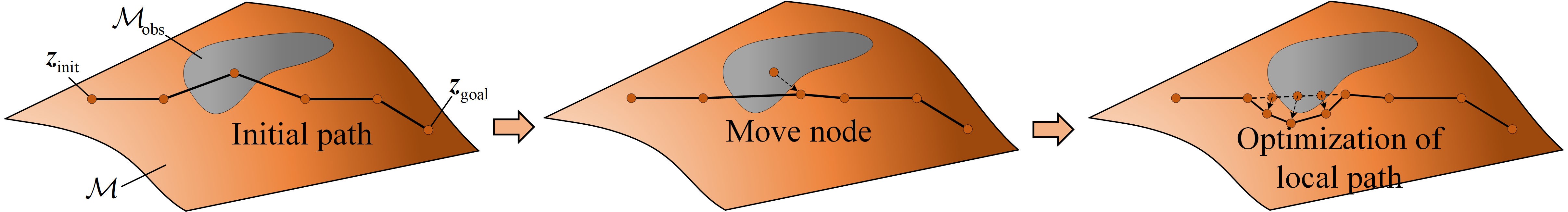} 
  \caption{Schematic of the local path optimization process in the latent space.}
  \label{fig3}
  \label{figurelabel}
\end{figure*}

In the motion planning task in this paper, the configuration space (C-space) of a robot with $n$ degrees of freedom is denoted as ${\cal Q} \in {\mathbb{R}^n}$,  comprising obstacle ${{\cal Q}_{{\rm{obs}}}}$ and obstacle-free ${{\cal Q}_{{\rm{free}}}} = {\cal Q}\backslash {{\cal Q}_{{\rm{obs}}}}$. The motion constraints involved in this paper can be expressed in terms of constraint function $h\left( \boldsymbol{q} \right) = {\bf{0}}$, $h(\boldsymbol{q}):{\mathbb{R}^n} \to {\mathbb{R}^l}$. All configurations satisfying the constraint function form the manifold: ${\cal M} = \left\{ {\boldsymbol{q} \in {\cal Q}|h(\boldsymbol{q}) = {\bf{0}}} \right\}$, $l$ denotes the number of constraints. When a configuration does not satisfy the constraint, the Jacobian matrix of the constraint function can be used to project the configuration onto the constraint manifold:

\begin{equation}
\boldsymbol{q} \leftarrow \boldsymbol{q} - \lambda {J_h}{(\boldsymbol{q})^\dag }h(\boldsymbol{q})
\label{eq:1}
\end{equation}

Where ${J_h}(\boldsymbol{q}){\rm{ = }}\partial h(\boldsymbol{q})/\partial \boldsymbol{q}$ denotes the Jacobian matrix of the constraint function and $\lambda $ denotes the step size parameter of the projection. If the value of $h(\boldsymbol{q})$ is less than the threshold $\varepsilon $ after a certain number of steps, the projection is considered successful. The process of projection is denoted by proj(). In actual tasks, there will be a series of similar motion constraints, and we use the constraint parameter $\boldsymbol{c}$ to represent different motion constraints. For example, in the task of wiping a table, the robot needs to constrain the movement on the surface of the table. We can use the height and posture of the table surface as task parameters to obtain the constraint function. The manifold also comprising obstacle ${{\cal M}_{{\rm{obs}}}} = {\cal M} \cap {{\cal Q}_{{\rm{obs}}}}$ and obstacle-free ${{\cal M}_{{\rm{free}}}} = {\cal M} \cap {{\cal Q}_{{\rm{free}}}}$. The constrained motion planning problem is to plan a collision-free motion path $\sigma$ that satisfies the constraints, given starting configuration ${\boldsymbol{q}_{{\rm{init}}}}$ and goal constraints ${{\cal Q}_{{\rm{goal}}}}$, such that $\sigma :[0,1] \to {{\cal M}_{{\rm{free}}}}$,$\sigma (0) = {\boldsymbol{q}_{{\rm{init}}}}$, $\sigma (1) \in {{\cal Q}_{{\rm{goal}}}}$.

\subsection{Constrained Motion Planning in Latent Space}

The constraint manifold is a space embedded in the configuration space with $n-l$ dimensions. Any configuration on the constraint manifold can be represented by a low-dimensional latent vector $\boldsymbol{z} \in {\mathbb{R}^{n - l}}$. In this paper, we use a Conditional Variational Autoencoders (CVAE) to approximate the manifold data into a low-dimensional latent space ${\cal Z} \in {\mathbb{R}^{n - l}}$. The mapping of the configuration space and the latent space to each other is realized by encoding neural network ${E_\phi }\left( \boldsymbol{q} \right):{\mathbb{R}^n} \to {\mathbb{R}^{n - l}}$ and decoding neural network ${D_\theta }\left( \boldsymbol{z} \right):{\mathbb{R}^{n - l}} \to {\mathbb{R}^n}$. The latent space can be divided into obstacle ${{\mathcal Z}_{{\rm{obs}}}} = {\rm{\{ }}\boldsymbol{z} \in {\cal Z}|{\rm{proj}}\left( {{D_\theta }\left( \boldsymbol{z} \right)} \right) \in {{\cal M}_{{\rm{obs}}}}{\rm{\} }}$ and obstacle-free ${{\cal Z}_{{\rm{free}}}} = {\rm{\{ }}\boldsymbol{z} \in {\cal Z}|{\rm{proj}}\left( {{D_\theta }\left( \boldsymbol{z} \right)} \right) \in {{\cal M}_{{\rm{free}}}}{\rm{\} }}$. Constrained motion planning in latent space is to plan a collision-free latent path ${\sigma _{\cal Z}}$,  given a starting latent vector ${\boldsymbol{z}_{{\rm{init}}}}\boldsymbol{ = }{E_\phi }\left( {{\boldsymbol{q}_{{\rm{init}}}}} \right)$  and a goal region ${{\cal Z}_{{\rm{goal}}}}{\rm{ = \{ }}\boldsymbol{z} \in {\cal Z}|{\rm{proj}}\left( {{D_\theta }\left( \boldsymbol{z} \right)} \right) \in {{\cal Q}_{{\rm{goal}}}}{\rm{\} }}$, such that ${\sigma _{\cal Z}}:[0,1] \to {{\cal Z}_{{\rm{free}}}}$, ${\sigma _{\cal Z}}(0) = {\boldsymbol{z}_{{\rm{init}}}}$, ${\sigma _{\cal Z}}(1) \in {{\cal Z}_{{\rm{goal}}}}$. Since the latent space can be continuously interpolated, the latent paths can be planned quickly without time-consuming projection.

The latest motion planning algorithm in latent space is Latent Constrained Bi-RRT(LCBiRRT) \cite{c13}, as shown in Algorithm ~\ref{algorithm1}. In LCBiRRT algorithm, a neural network is used to directly check the validity of waypoint in latent path. Since the validity checking neural network has errors, it is necessary to map the latent path into the configuration space to further check its validity. After mapping to the configuration space, if the latent waypoint cannot be projected to the constraint manifold or collides with the obstacle, the waypoint are deleted from the tree and the latent path is replanned. In this paper, the path validity check algorithm is improved to reduce the planning time.

\begin{algorithm}
\caption{Latent Constrained Bi-RRT}
\label{algorithm1}

${\boldsymbol{z}_{{\rm{init}}}} \leftarrow {E_{\Phi }}\left( {{\boldsymbol{q}_{{\rm{init}}}}} \right);{\boldsymbol{z}_{{\rm{goal}}}} \leftarrow {E_{\Phi }}\left( {{\boldsymbol{q}_{{\rm{goal}}}}} \right)$

${\boldsymbol{x}_{{\rm{init}}}} \leftarrow \left( {{\boldsymbol{q}_{{\rm{init}}}},{\boldsymbol{z}_{{\rm{init}}}}} \right);{\boldsymbol{x}_{{\rm{goal}}}} \leftarrow {E_{\Phi }}\left( {{\boldsymbol{q}_{{\rm{goal}}}},{\boldsymbol{z}_{{\rm{goal}}}}} \right)$

${{\cal T}_a}.{\rm{init}}({\boldsymbol{x}_{{\rm{init}}}});{{\cal T}_b}.{\rm{init}}({\boldsymbol{x}_{{\rm{goal}}}});n \leftarrow 0$

\While{not timeout}{
    SampleStartGoal()
    
    ${\boldsymbol{z}_{{\rm{rand}}}} \leftarrow $ RandomConfigZ()
    
    ${\boldsymbol{x}_a},{\rm{res}} \leftarrow $ ConstrainedExtendZ $\left( {{{\cal T}_a},{\boldsymbol{z}_{{\rm{rand}}}}} \right)$

    \If{\rm{res} $ \ne $ \textit{Trapped}}{
        ${\boldsymbol{x}_{{\rm{new}}}},{\rm{res}} \leftarrow $ ConstrainedExtendZ $\left( {{\cal T}_b},{\boldsymbol{z}_a} \right) $
        
        \While{\rm{res}=\textit{Advanced}}{
            ${\boldsymbol{x}_{{\rm{new}}}},{\rm{res}} \leftarrow $ ConstrainedExtendZ $\left( {{\cal T}_b},{\boldsymbol{z}_a} \right) $
        }
        
        \If{\rm{res}=\textit{Reached}}{
            $n \leftarrow n + 1$ 

            \If{\textcolor{red}{CheckPathValid$\left( {{\cal T}_a},{\boldsymbol{x}_a},n \right)$} =False }{
                ${\boldsymbol{x}_{{\rm{last}}}} \leftarrow $ GetLastValid $\left( {{\cal T}_a},{\boldsymbol{x}_a} \right) $

                ${\rm{res}} \leftarrow $ LatentJump $\left( {{\cal T}_a},{{\cal T}_b},{\boldsymbol{q}_{{\rm{last}}}} \right) $

                \If{\rm{res} $ \ne $ \textit{Trapped}}{ \textbf{continue} }
            }
            
            \If{ \textcolor{red}{CheckPathValid$\left( {{{\cal T}_b},{\boldsymbol{x}_{{\rm{new}}}},n} \right)$} =False }{
                ${\boldsymbol{x}_{{\rm{last}}}} \leftarrow $ GetLastValid $\left( {{\cal T}_b},{\boldsymbol{x}_{{\rm{new}}}} \right) $

                ${\rm{res}} \leftarrow $ LatentJump $\left( {{\cal T}_b},{{\cal T}_a},{\boldsymbol{q}_{{\rm{last}}}} \right) $

                \If{\rm{res} $ \ne $ \textit{Trapped}}{ \textbf{continue} }
            }

            \textbf{return} path$\left( {{\cal T}_a},{{\cal T}_b} \right) $
        }
    }
    \If{${\boldsymbol{p}_q} >$ \rm{sample from} $ U\left( {0,1} \right)$}{
    
        ${\boldsymbol{q}_{{\rm{rand}}}} \leftarrow $ RandomConfigQ()

        ConstrainedExtendQ$\left( {{\cal T}_a},{\boldsymbol{q}_{{\rm{rand}}}} \right) $
        
    }

    Swap$\left( {{\cal T}_a},{{\cal T}_b} \right) $
    
}

\end{algorithm}

\section{METHED}

\subsection{Overview}

\begin{algorithm}
\caption{CheckPathValid(${\cal T},{\boldsymbol{x}_{{\rm{dst}}}},n$)}\label{algorithm2}

$\boldsymbol{X} \leftarrow$ GetPath(${\cal T},{\boldsymbol{x}_{{\rm{dst}}}}$)

\eIf{$n$ \rm{is divisible by} $interva$l}{
    \For{$\boldsymbol{x}_i \in \boldsymbol{X}$}{
        ${\boldsymbol{q}_{{\rm{proj}}}} \leftarrow \rm{Project} \left( {D_\theta }(\boldsymbol{z}_\textit{i}) \right)$

        \If{\rm{IsValid}${\left( \boldsymbol{q}_{{\rm{proj}}} \right)}$= False}{
            ${\boldsymbol{q}_{{\rm{move}}}},{\boldsymbol{z}_{{\rm{move}}}} \leftarrow $MoveNodeZ($\boldsymbol{z}_i$)

            \If{${\boldsymbol{q}_{{\rm{move}}}}$=\rm{NULL}}{
                ${\cal T}$.DeleteBranch($\boldsymbol{x}_i$)

                \textbf{return} False
            }

            ${\boldsymbol{q}_{{\rm{proj}}}} \leftarrow {\boldsymbol{q}_{{\rm{move}}}}$
        }

        \If{\rm{IsValid}${\left( {\boldsymbol{q}_{{i\rm{-1}}}},{\boldsymbol{q}_{{\rm{proj}}}} \right)}$= False}{

            $\boldsymbol{Q} \leftarrow $\rm{InterpolateQ}${\left( {\boldsymbol{q}_{{i\rm{-1}}}},{\boldsymbol{q}_{{\rm{proj}}}} \right)}$

            \For{$\boldsymbol{q}_j \in \boldsymbol{Q}$}{
                ${\boldsymbol{q}_{{\rm{move}}}},{\boldsymbol{z}_{{\rm{move}}}} \leftarrow $MoveNodeZ$\left( {E_{\Phi }}\left( \boldsymbol{q}_j \right) \right)$

                \If{${\boldsymbol{q}_{{\rm{move}}}}$=\rm{NULL} or \rm{IsValid}${\left( {\boldsymbol{q}_{{j\rm{-1}}}},{\boldsymbol{q}_{{\rm{move}}}} \right)}$= False}{
                ${\cal T}$.DeleteBranch($\boldsymbol{x}_i$)

                \textbf{return} False
            }
            }
        
        }
        $\boldsymbol{x}_i \leftarrow ({\boldsymbol{q}_{{\rm{proj}}}},\boldsymbol{z}_i)$
    }
}{
    \For{$\boldsymbol{x}_i \in \boldsymbol{X}$}{
        ${\boldsymbol{q}_{{\rm{proj}}}} \leftarrow \rm{Project} \left( {D_\theta }(\boldsymbol{z}_i) \right)$
        
        \If{\rm{IsValid}${\left( \boldsymbol{q}_{{\rm{proj}}} \right)}$= False or \rm{IsValid}${\left( {\boldsymbol{q}_{{i\rm{-1}}}},{\boldsymbol{q}_{{\rm{proj}}}} \right)}$= False}{
            ${\cal T}$.DeleteBranch($\boldsymbol{x}_i$)

            \textbf{return} False
        }
        $\boldsymbol{x}_i \leftarrow ({\boldsymbol{q}_{{\rm{proj}}}},\boldsymbol{z}_i)$
        }

}
\textbf{return} True
\end{algorithm}

\begin{algorithm}
\caption{MoveNodeZ($\boldsymbol{z}$)}\label{algorithm3}
\For{$i \leftarrow 0$ \KwTo  ${N_{\max }}$}{
    $\boldsymbol{z} \leftarrow \boldsymbol{z} + \gamma {\rm{ }}{\nabla _{\boldsymbol{z}}}{P_\psi }\left( {{\boldsymbol{z}_S},\boldsymbol{c},\boldsymbol{z}} \right)$

    ${\boldsymbol{q}_{{\rm{proj}}}} \leftarrow \rm{Project} \left( {D_\theta }(\boldsymbol{z}) \right)$

    \If{\rm{IsValid}${\left( \boldsymbol{q}_{{\rm{proj}}} \right)}$= True}{
        \textbf{return} ${\boldsymbol{q}_{{\rm{proj}}}},\boldsymbol{z}$
    }
}

\textbf{return} NULL, NULL

\end{algorithm}

The workflow of the proposed method is shown in Fig.~\ref{fig2}, which includes five neural networks, namely, configuration encoder, voxel encoder, Signed Distance Field (SDF) encoder, validity check network, and minimum distance prediction network. Except SDF encoder and minimum distance prediction network, other neural networks are the same as in \cite{c13}. The configuration encoder ${E_\phi }\left( \boldsymbol{q} \right)$ is used to map the configuration $\boldsymbol{q}$ on the constraint manifold into the latent space ${\cal Z}$. CVAE is used to train the configuration encoder ${E_\phi }\left( \boldsymbol{q} \right)$ and decoder ${D_\theta }\left( \boldsymbol{z} \right)$ simultaneously. Generalization over different constraints is achieved by training with a randomly selected constraint parameter $\boldsymbol{c}$.

The voxel encoder ${E_\varphi }\left( \boldsymbol{V} \right) \to {\boldsymbol{z}_V} \in {\mathbb{R}^{{d_1}}}$ and SDF encoder ${E_\omega }\left( \boldsymbol{S} \right) \to {\boldsymbol{z}_S} \in {\mathbb{R}^{{d_2}}}$ are used to extract the features of the environment voxel $\boldsymbol{V}$ and the environment SDF $\boldsymbol{S}$, respectively, and the environment SDF is calculated based on voxel. The validity check network ${V_\xi }\left( {{\boldsymbol{z}_V},\boldsymbol{c},{\boldsymbol{z}_t}} \right) \to [0,1]$ takes the latent vector ${\boldsymbol{z}_t}$, the features of the voxel grid ${\boldsymbol{z}_V}$ and the constraint parameter $\boldsymbol{c}$ as input to check the validity of the current latent vector ${\boldsymbol{z}_t}$. If ${D_\theta }\left( {{\boldsymbol{z}_t}} \right)$ fails to project onto the constraint manifold or collision with the obstacle, it is considered invalid. The voxel encoder and the validity check network are trained simultaneously, and the cross-entropy loss function is used.

The minimum distance prediction network ${P_\psi }\left( {{\boldsymbol{z}_S},\boldsymbol{c},{\boldsymbol{z}_i}} \right) \to {\hat d_{\min }}$ takes the latent waypoint ${\boldsymbol{z}_i}$, the SDF features ${\boldsymbol{z}_S}$ and the constraint parameter $\boldsymbol{c}$ as input, and predicts the minimum distance between the robot and the obstacles in the task space. The SDF encoder and the minimum distance prediction network were trained together, and the Mean Square Error (MSE) loss function was used. To calculate the true value of ${d_{\min }}$, ${\boldsymbol{z}_i}$ is first mapped to the corresponding constrained configuration ${\boldsymbol{q}_i} = {\rm{proj}}\left( {{D_\theta }\left( \boldsymbol{z}_i \right)} \right)$, and then the forward kinematics of the robot is used to calculate the positions of spheres for enveloding the robot. Finally, the minimum distance is obtained by using SDF.

The LCBiRRT algorithm takes the starting latent vector ${\boldsymbol{z}_s}$, the target latent vector ${\boldsymbol{z}_g}$ and the constraint parameters $\boldsymbol{c}$ as input, randomly expands the nodes in the latent space, and uses the validity check network to test the validity of the latent vector. When a feasible latent path is found, the path validity check algorithm is used to verify the path. In order to reduce the planning time, we use the minimum distance prediction network to locally optimize the latent space path in the path validity check algorithm.

\subsection{Local Path Optimization Algorithm}

The path ${\boldsymbol{P}_{\boldsymbol{z}}} = \{ {\boldsymbol{z}_i}\} _{i = 1}^k$ obtained by planning in the latent space still has the possibility of entering the obstacle area. In order to reduce the time of replanning, this paper proposes a local path optimization method in latent space, as shown in Algorithm ~\ref{algorithm2}. Firstly, the waypoints ${\boldsymbol{z}_i}$ in the latent space are decoded into the configuration space, ${\boldsymbol{\hat q}_i} = {D_\theta }({\boldsymbol{z}_i})$, and the projection method in (1) is used to project ${\boldsymbol{\hat q}_i}$ onto the constrained manifold, ${\boldsymbol{q}_i}{\rm{ = proj(}}{\boldsymbol{\hat q}_i}{\rm{)}}$. If the projection is successful and ${\boldsymbol{q}_i} \in {{\cal Q}_{{\rm{obs}}}}$, then try to move the latent waypoint away from the obstacle area in the latent space. If the waypoint $\boldsymbol{z}_i$ moves successfully, the validity of the local path between waypoints is detected. Otherwise, the collision waypoints are deleted and the latent path is replanned.

If the local path between waypoints enters the obstacle area, start to optimize the local path for obstacle avoidance. Firstly, the configuration of adjacent waypoints is interpolated on the constrained manifold to obtain the local waypoints $\boldsymbol{Q}$. Then, the configuration in the local waypoints is encoded into the latent space, and the latent waypoints is moved away from the obstacle, then the validity between the local waypoints is checked. If the local path optimization fails, the collision nodes is deleted and the path is replanned. Because the local path optimization is time consuming, the local path optimization is carried out at a certain interval in the algorithm to reduce the overall planning time. The schematic of the local path optimization is shown in Fig.~\ref{fig3}.

\subsection{Obstacle Avoidance Movement in Latent Space}

When the latent vector is in an obstacle area, i.e. $\boldsymbol{z} \in {{\cal Z}_{{\rm{obs}}}}$, we want to move the latent vector away from the obstacle area. The moving direction of the latent vector can be estimated by using the gradient of the minimum distance prediction neural network, and the latent vector can gradually move away from the obstacle by gradient ascent:

\begin{equation}
\frac{{\partial {{\hat d}_{\min }}}}{{\partial \boldsymbol{z}}} \approx {\nabla _{\boldsymbol{z}}}{P_\psi }\left( {{\boldsymbol{z}_S},\boldsymbol{c},\boldsymbol{z}} \right)
\label{eq:2}
\end{equation}

\begin{equation}
\boldsymbol{z} \leftarrow \boldsymbol{z} + \gamma {\rm{ }}{\nabla _{\boldsymbol{z}}}{P_\psi }\left( {{\boldsymbol{z}_S},\boldsymbol{c},\boldsymbol{z}} \right)
\label{eq:3}
\end{equation}

Where $\gamma  \in {{\rm{R}}^{\rm{ + }}}$ is a hyperparameter representing the step size. The algorithm for moving the latent vectors is shown in Algorithm ~\ref{algorithm3}. After each movement of the latent vector, it is first decoded into the configuration space $\boldsymbol{\hat q} = {D_\theta }(\boldsymbol{z})$. Then the projection method in (1) is used to project $\boldsymbol{\hat q}$ onto the constraint manifold and perform collision detection for the robot. When the robot no longer collides with the obstacle, the latent vector stops moving.

\begin{figure*}[thpb]
  \centering
  \subfigbottomskip=2pt 
  \subfigcapskip=-2pt 
  \subfigure[The path of the robot in Scenario 1.]{
     \label{fig4-a}
     \includegraphics[width=0.95\linewidth]{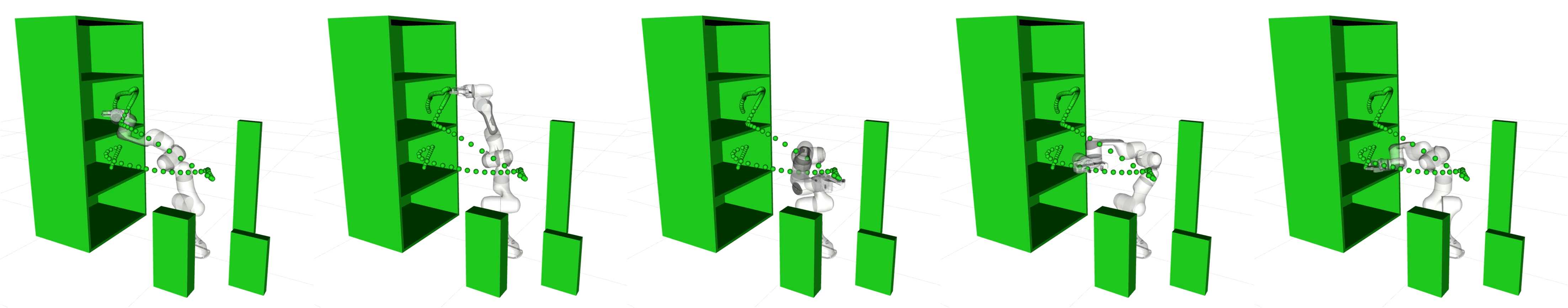}}
  \subfigure[The path of the robot in Scenario 2.]{
     \label{fig4-b}
     \includegraphics[width=0.95\linewidth]{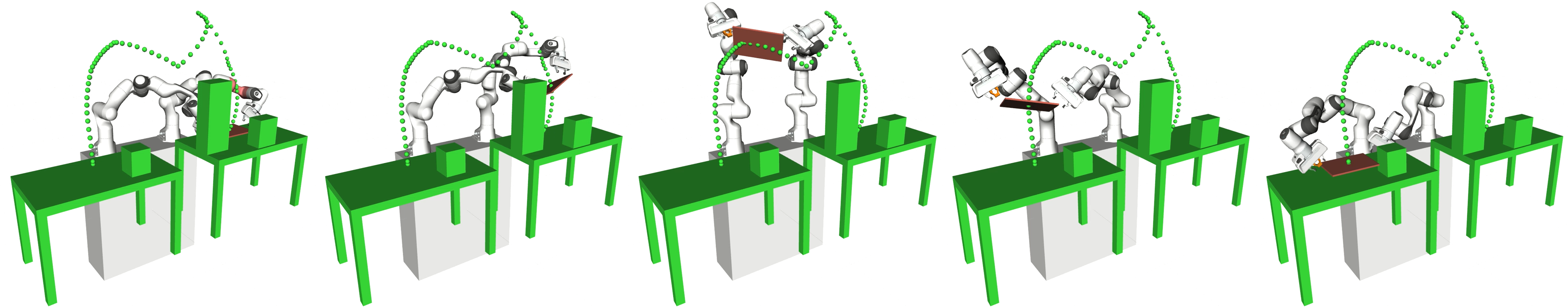}}
  \subfigure[The path of the robot in Scenario 3.]{
     \label{fig4-c}
     \includegraphics[width=0.95\linewidth]{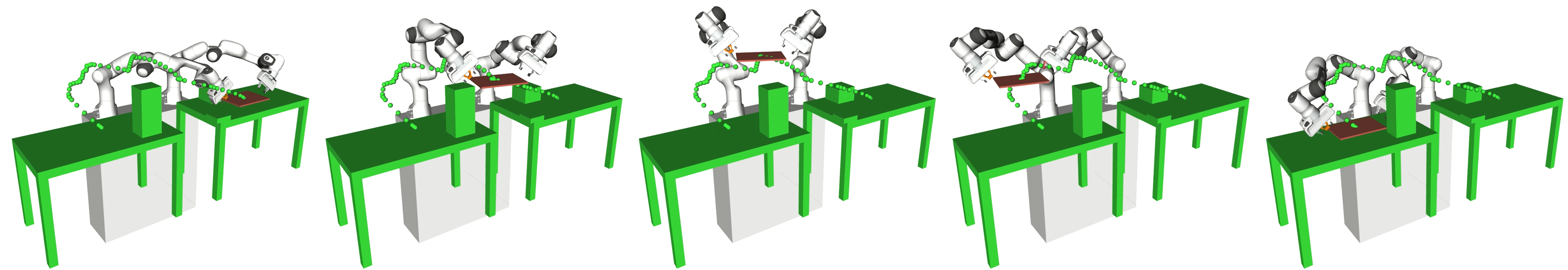}}
  \caption{Three experimental scenarios and the robot motion paths.}
  \label{fig4}
  \label{figurelabel}
\end{figure*}

\begin{figure}
  \centering
  \includegraphics[width=0.4\textwidth]{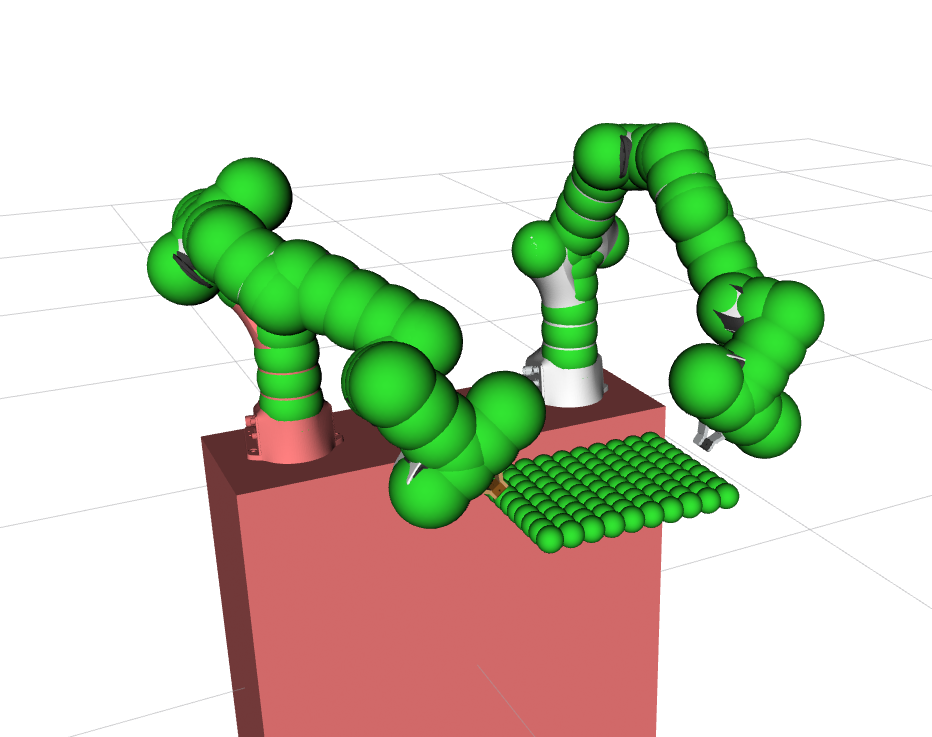} 
  \caption{The spheres used to envelope the robot.}
  \label{fig5}
  \label{figurelabel}
\end{figure}

\begin{figure}
  \centering
  \subfigbottomskip=2pt 
  \subfigcapskip=-2pt 
  \subfigure[Mean path search time and mean path check time for Scenario 2.]{
     \label{fig6-a}
     \includegraphics[width=0.9\linewidth]{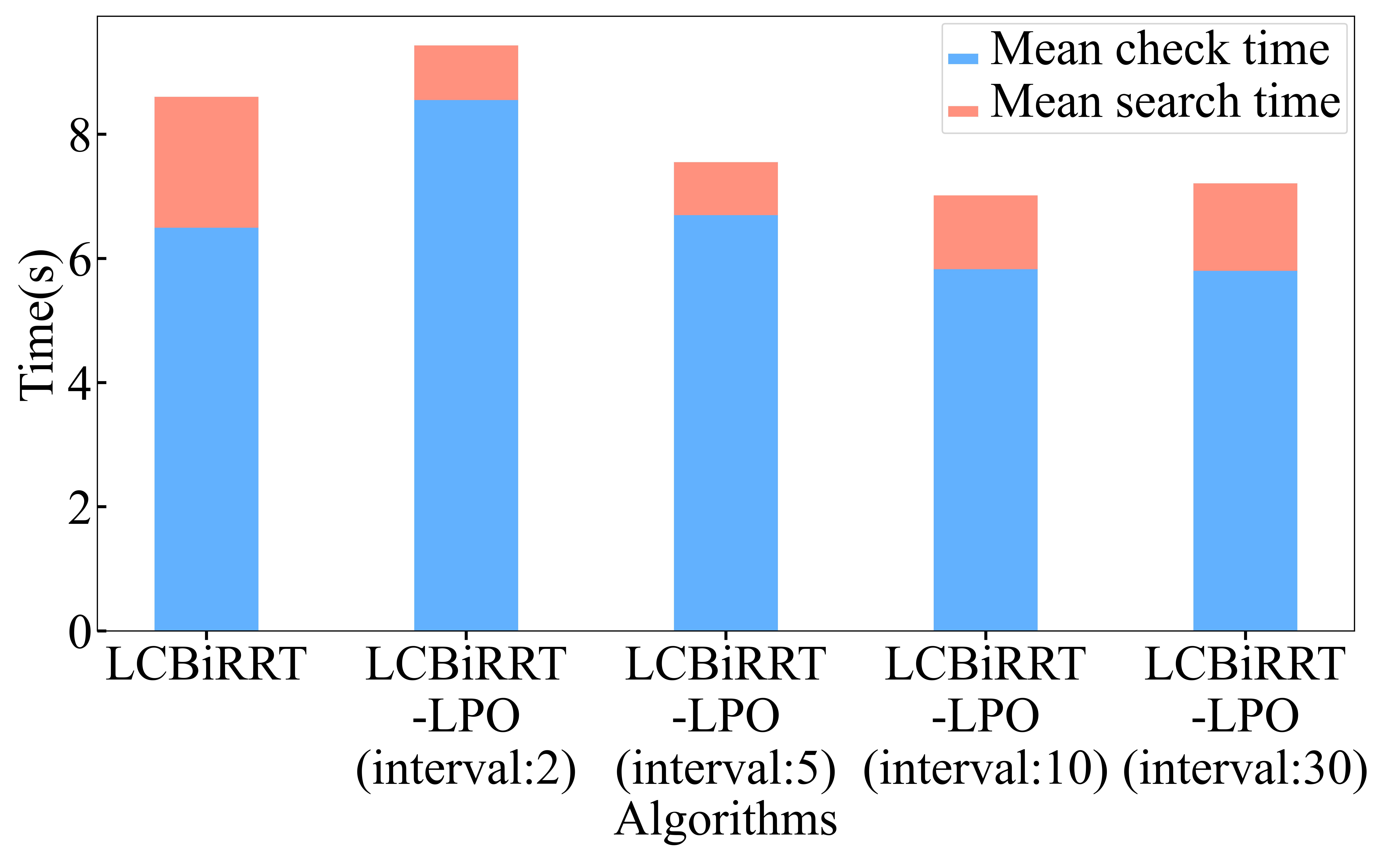}}
  \subfigure[Mean path search time and mean path check time for Scenario 3.]{
     \label{fig6-b}
     \includegraphics[width=0.9\linewidth]{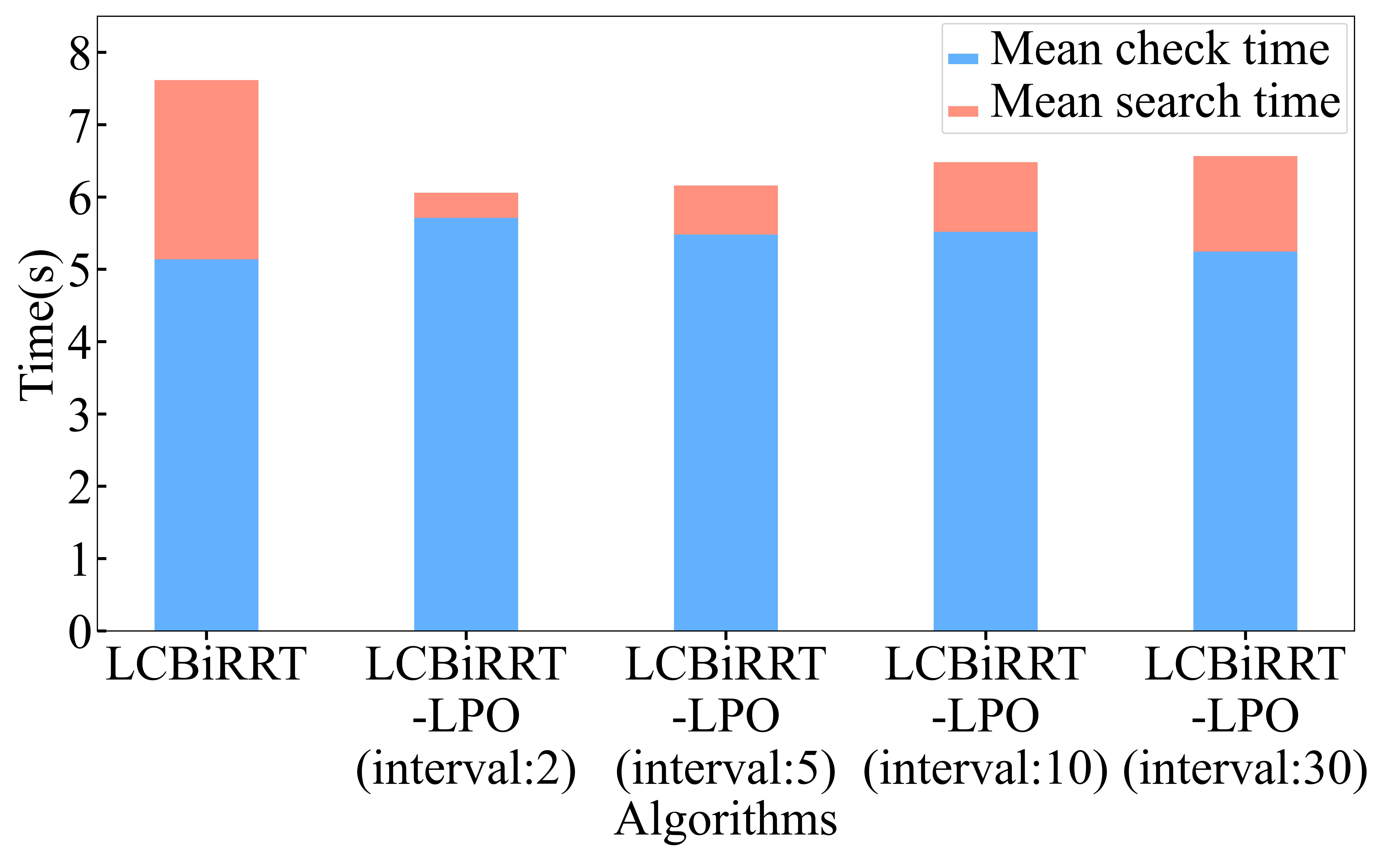}}
  \caption{Test results for mean path search time and mean path check time.}
  \label{fig6}
  \label{figurelabel}
\end{figure}

\section{EXPERIMENTS}

\subsection{Experiment Setup} 
Four constrained motion planning algorithms, CBiRRT2 \cite{c6}, Latent Sampling \cite{c12}, Precomputed Graph \cite{c9} and LCBiRRT \cite{c13}, were selected for comparison. Based on the LCBiRRT algorithm, Local Path Optimization(LOP) with different intervals(interval: 2, 5, 10, 30) are used for testing. In this paper, all algorithms are tested in three scenarios \cite{c13}, as shown in Fig.~\ref{fig4}. In scenario 1, fixed orientation constraints are applied to a single Franka Panda manipulator, the degree of freedom of the system is 7 and the constraint dimension is 2. In scenario 2, closed-chain constraints are applied to two Franka Panda manipulators, the degree of freedom of the system is 14, and the constraint dimension is 6. In scenario 3, the closed-chain constraint and fixed orientation constraint are applied to two Franka Panda manipulators. The degree of freedom of the system is 14 and the dimension of the constraint is 8. In scenario 2 and scenario 3, the condition parameters $\boldsymbol{c}$ include the tray length, handle length, and handle angle. The condition ranges for these parameters are 0.2–0.6 m, 0.02–0.1 m, and $0^\circ $ to $90^\circ $, respectively. The step size $\gamma$ is set to 0.8, the moving steps ${N_{\max }}$ is set to 10.

100 planning experiments were performed in each scenario, and the starting configuration, the goal configuration, and the position and size of obstacles were randomly selected in advance. The kinematic model of the robot and the collision detection program are implemented in the MoveIt library, and the motion planning algorithms were implemented in Python. The program runs on an Nvidia Geforce RTX3090 GPU and an intel i9-10900K CPU with a planned time limit of 300s.

\begin{table}
\centering
\setlength{\abovecaptionskip}{0.cm}
\caption{The Expermental Results of Scenario 1}
\label{table:1}
\begin{tabular}{c c c} 
\toprule 
Algorithms & Time & Success rate\\
\midrule 

CBiRRT2	& 15.771±17.111	& 1.0 \\
Precomputed graph 	& 18.033±33.824	& 1.0\\
CBiRRT2 with latent sampling	& 15.943±27.040	& 1.0\\
LCBiRRT	& 2.242±4.270	& 1.0\\
LCBiRRT-LPO (interval:2)	& 2.974±4.069	& 1.0\\
LCBiRRT-LPO (interval:5)	& 2.134±2.829	& 1.0\\
LCBiRRT-LPO (interval:10)	& 2.517±3.548 & 1.0\\
LCBiRRT-LPO (interval:30)	& \textbf{2.052±2.639} &	1.0\\

\bottomrule 
\end{tabular}
\end{table}

\begin{table}
\centering
\setlength{\abovecaptionskip}{0.cm}
\caption{The Expermental Results of Scenario 2}
\label{table:2}
\begin{tabular}{c c c} 
\toprule 
Algorithms & Time & Success rate\\
\midrule 

CBiRRT2	& 97.276±92.451	& 0.71\\
Precomputed graph 	& 91.382±92.853	& 0.78\\
CBiRRT2 with latent sampling	& 85.503±84.551	& 0.87\\
LCBiRRT	& 8.601±16.754	& 1.0\\
LCBiRRT-LPO (interval:2)	& 9.430±12.419	& 1.0\\
LCBiRRT-LPO (interval:5)	& 7.550±12.048	& 1.0\\
LCBiRRT-LPO (interval:10)	& \textbf{7.013±8.856}	& 1.0\\
LCBiRRT-LPO (interval:30)	& 7.207±9.362	& 1.0\\

\bottomrule 
\end{tabular}
\end{table}

\begin{table}[t]
\centering
\setlength{\abovecaptionskip}{0.cm}
\caption{The Expermental Results of Scenario 3}
\label{table:3}
\begin{tabular}{c c c} 
\toprule 
Algorithms & Time & Success rate\\
\midrule 

CBiRRT2	&84.646±90.431	&0.76\\
Precomputed graph 	&68.609±73.442	&0.86\\
CBiRRT2 with latent sampling	&52.388±63.439	&0.95\\
LCBiRRT	&7.615±11.768	&1.0\\
LCBiRRT-LPO (interval:2)	& \textbf{6.059±8.173}	&1.0\\
LCBiRRT-LPO (interval:5)	&6.159±6.489	&1.0\\
LCBiRRT-LPO (interval:10)	&6.480±10.450	&1.0\\
LCBiRRT-LPO (interval:30)	&6.564±8.093	&1.0\\

\bottomrule 
\end{tabular}
\end{table}

\subsection{Dataset and Training Details}

In each experimental scenario, 10000 on-manifold configurations are generated for the training of CVAE. By randomly changing the position of the obstacles, 500 voxels is generated in each scenario for training the voxel encoder and the validity check network, and the SDF corresponding to the voxels are calculated. The size of both the voxel grid and SDF is ${\rm{32}} \times {\rm{32}} \times {\rm{32}}$. The training details of configuration encoder, voxel encoder and validity check network are the same as \cite{c13}. 

In each SDF, 200 on-manifold configurations are randomly generated, of which 100 configurations do not collide with obstacles and 100 configurations collide with obstacles. The constraint parameter $\boldsymbol{c}$ is randomly selected in the process of computing the configurations. The minimum distance corresponding to each configuration is calculated using the SDF and the position of the envelope spheres. The envelope spheres is shown in the Fig.~\ref{fig5}. Finally, 100,000 minimum distance data are generated for each scenario, where 90\% of the data was used as the training set and 10\% of the data was used as the test set. SDF encoder and voxel encoder adopt the same structure, including two hidden layers with 512 nodes. The input of the SDF encoder is one-dimensionalized SDF data, the output is 16-dimensional features, and the activation function is Leaky Rectified Linear Unit (Leaky ReLU). The distance prediction neural network consists of three hidden layers with a number of 1024 nodes, and the activation function is Rectified Linear Unit (ReLU). SDF encoder and minimum distance prediction neural network are trained together, the learning rate is 0.003, the batch size is 64, and the number of epoch is 60. The code of the algorithms is available at \href{https://github.com/hit618/LCBiRRT-LPO}{https://github.com/hit618/LCBiRRT-LPO}.

\subsection{Motion Planning Results}

The experimental results of motion planning are shown in TABLE~\ref{table:1} -~\ref{table:3}. The mean and standard deviation of the planning time are calculated, the success rate indicates the ratio of solved problems within the time limit. The proposed method achieves the best performance in terms of success rate and planning time. Compared to the state-of-the-art algorithm LCBiRRT, our method has a greater advantage in complex planning scenarios (Scenario 2 and 3) and achieves similar results to LCBiRRT in simple planning scenarios (Scenario 1). This is because more path search time and replanning times are required in complex scenarios, while the proposed method can effectively reduce the above time-consuming.

The mean path search time and mean path check time for Scenario 2 and Scenario 3 is counted, as shown in Fig.~\ref{fig6}. As the interval increases, the path search time increases, while the path checking time decreases. This shows that the proposed method has a significant effect in reducing the number of path replanning and reducing the path search time. Because the process of local path optimization is time-consuming, too frequent local path optimization may lead to an increase in the path check time. Therefore, suitable optimization interval should be selected to achieve the overall best planning time.


\section{CONCLUSIONS}

This paper proposes a local path optimization algorithm in latent space for improving the speed of constrained motion planning. By training a neural network to predict the minimum distance between the robot and the obstacle with the latent vector as input, the learned distance gradient is used to guide the movement of the waypoints in the latent space, which successfully reduces the planning time. Compared with the state-of-the-art algorithms in three planning scenarios with different difficulty levels, the proposed method achieves the fastest planning speed. In future research, a global path optimization algorithm will use the learned distance gradient to improve path quality. In addition, because the path validity check process is time-consuming, faster path check algorithms need to be further studied.

\addtolength{\textheight}{-5cm}   




\end{document}